\documentclass[runningheads,a4paper]{llncs}
\usepackage{graphicx}
\usepackage{amsmath}
\usepackage{amssymb}
\usepackage[utf8]{inputenc}
\usepackage{url}
\usepackage{color}

\fontfamily{ptm}
\begin{document}
\frontmatter          

\pagenumbering{arabic}
\pagestyle{headings}  

\title{TRANSALG: a Tool for Translating Procedural Descriptions of Discrete Functions to SAT}
\titlerunning{TRANSALG : translating discrete functions to SAT}  
\author{Ilya Otpuschennikov, Alexander Semenov \and Stepan Kochemazov}
\authorrunning{I.\,Otpuschennikov, A.\,Semenov \and S.\,Kochemazov}
\institute{Matrosov Institute for System Dynamics and Control Theory of 
Siberian Branch of Russian Academy of Sciences
(ISDCT SB RAS), Irkutsk, Russia
}

\maketitle              

\begin{abstract}
In this paper we present the \textsc{Transalg} system, designed to produce SAT encodings for discrete functions, written as programs in a specific language. Translation of such programs to SAT is based on propositional encoding methods for formal computing models and on the concept of symbolic execution. We used the \textsc{Transalg} system to make SAT encodings for a number of cryptographic functions.
\end{abstract}
\section{Introduction}
Many new methods for solving Boolean Satisfiability Problem (SAT) were introduced in the past two decades. These methods make it possible to solve combinatorial problems from various areas \cite{DBLP:series/faia/2009-185}. One can use different approaches to encode an original problem to SAT \cite{DBLP:series/faia/Prestwich09}. Often each particular problem requires researchers to develop and implement special encoding technique. Recently a number of systems that automate procedures of encoding combinatorial problems to SAT were developed \cite{DBLP:conf/cpaior/HebrardOO10,journals/lmcs/predrag,DBLP:journals/tplp/MetodiC12,DBLP:conf/sat/SohTB13,tamura2008system}.

In our paper, we present the \textsc{Transalg} system that translates procedural descriptions of discrete functions to SAT. Translation mechanisms employed in \textsc{Transalg} are based on the ideas of S.A.\,Cook on propositional encoding of Turing Machine programs \cite{Cook:1971:CTP:800157.805047} and ideas of J.S.\,King on symbolic execution \cite{King:1976:SEP:360248.360252}. At this time we mainly use \textsc{Transalg} to produce SAT encodings for cryptographic functions. Inversion problems for such functions are usually computationally hard. Recent works show that the study of cryptanalysis problems in the context of SAT approach can yield promising results \cite{DBLP:conf/sat/EibachPV08,Mcdonald_attackingbivium,DBLP:conf/sat/MironovZ06,DBLP:conf/pact/SemenovZBP11,DBLP:conf/tools/Soos10,DBLP:conf/sat/SoosNC09}. In our opinion \textsc{Transalg} might become a powerful tool in the research of various cryptographic functions.

To be processed by \textsc{Transalg} a function description should be written in the special TA-language. The TA-language is a procedural domain specific language (DSL) with C-like syntax. Therefore it is usually sufficient to introduce minor changes to C-implementation of an algorithm to produce its description in the TA-language.

It should be noted that the base concept of \textsc{Transalg} makes it possible to produce SAT encodings for arbitrary discrete functions computable in polynomial time. We believe that \textsc{Transalg} will be useful for many researchers who want to use SAT approach in their studies but don’t have time or desire to develop special encoding techniques for each individual problem.

\section{The TA-language for Description of Discrete Functions}

As we already mentioned above, the \textsc{Transalg} system uses special DSL, named TA-language, to describe discrete functions. Below by translation we mean the process of obtaining a propositional encoding for a TA-program. 

The translation of a TA-program has two main stages. At the first stage, \textsc{Transalg} parses a source code of a TA-program and constructs a syntax tree using standard techniques of compilation theory \cite{Aho:2006:CPT:1177220}. At the second stage, the system employs the concept of symbolic execution \cite{King:1976:SEP:360248.360252} to construct a propositional encoding for a TA-program considered. \textsc{Transalg} can output an encoding obtained in any of standard forms (CNF, DNF, ANF).

The TA-language has block structure. A block (composite operator) is a list of instructions. Every block is delimited by curly braces '\texttt{\{}' and '\texttt{\}}'. Within a block instructions are separated by '\texttt{;}' symbol. Each block has its own (local) scope. In the TA-language nested blocks are allowed with no limit on depth. During the analysis of a program \textsc{Transalg} constructs a scope tree with global scope at its root. Every identifier in a TA-program belongs to some scope. Variables and arrays declared outside of any block and also all functions belong to a global scope and therefore can be accessed in any point of a program.

A TA-program is essentially a list of functions. The \texttt{main} function is an entry point so it must exist in any program. The TA-language supports base constructions used in procedural languages (variable declarations, assignment operators, conditional operators, loops, function calls, function returns etc.), various integer operations and bit operations including bit shifting and comparison.

Main data type in the TA-language is the \texttt{bit} type. \textsc{Transalg} uses this type to establish links between variables used in a TA-program and Boolean variables included into corresponding propositional encoding. It is important to distinguish between these two sets of variables. Below we will refer to variables that appear in a TA-program as \textit{program variables}. All variables included in a propositional encoding are called \textit{encoding variables}. Upon the translation of an arbitrary instruction that contains a program variable of the \texttt{bit} type, \textsc{Transalg} links this program variable with a corresponding encoding variable. \textsc{Transalg} establishes such links only for program variables of the \texttt{bit} type. Variables of other types, in particular of the \texttt{int} type and the \texttt{void} type are used only as service variables, for example as loop counters or to specify functions without return value.

Declarations of global \texttt{bit} variables can have \texttt{\_\_in} or \texttt{\_\_out} attribute. Attribute \texttt{\_\_in} marks variables that contain input data for an algorithm. Attribute \texttt{\_\_out} marks variables that contain an output of an algorithm. Local \texttt{bit} variables cannot be declared with these attributes.

\section{Translation of TA-programs to SAT}

Let us consider a sequence of computations defined by an arbitrary TA-program as a sequence of data modifications in a memory of an abstract computing machine in moments $\{0,1,\ldots,e\}$. At every moment $i\in\{0,1,\ldots,e\}$ \textsc{Transalg} associates a set $X^i$ of encoding variables with program variables of the \texttt{bit} type. Denote $X=\bigcup_{i=0}^{e} X^i$. Suppose that $X^{in}$ is formed by encoding variables that correspond to input data, and $X^{out}$ contains encoding variables corresponding to the output of a discrete function considered. It is easy to see that $X^{in}\subseteq X^0$ and $X^{out}\subseteq X$.

\textsc{Transalg} uses the translation concept that makes it possible to reduce the redundancy of propositional encoding. We will explain this concept on the following example.
\begin{example}
Consider an encoding of a linear feedback shift register (LFSR) \cite{Menezes:1996:HAC:548089} with \textsc{Transalg}. In fig. \ref{rslos_ta} we show the TA-program for the LFSR with feedback polynomial $P(z)=z^{19}+z^{18}+z^{17}+z^{14}+1$ over $GF(2)$ (here $z$ is a formal variable).
\begin{figure}[htbp]
	\centering
		\includegraphics[width=12cm]{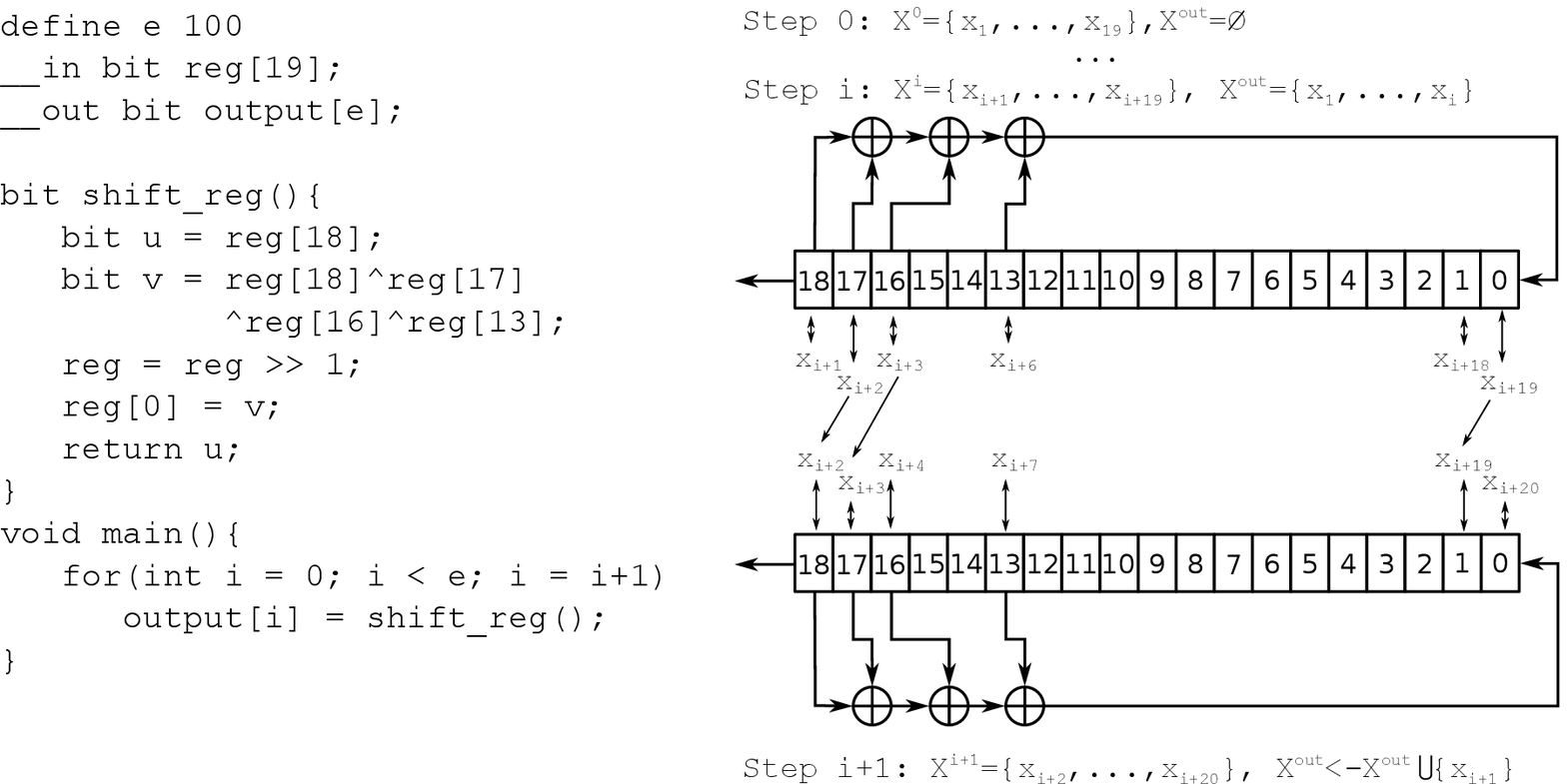}
		\caption{TA-program for the LFSR}
		\label{rslos_ta}
\end{figure}

Note that during the translation of transition from step $i$ to step $i+1$ it is not necessary to create new encoding variables for every cell of the register. If we copy data from one register cell to another, then we can use the same encoding variable to encode corresponding data value at steps $i$ and $i+1$. Therefore, at each step \textsc{Transalg} creates only one new encoding variable and links it with program variable \texttt{reg[0]}. All the other program variables get linked with encoding variables created at previous steps. In accordance with the above, a set of encoding variables corresponding to initial values of the register is $X^{in}=X^0=\{x_{1},x_{2},\ldots,x_{19}\}$. After each shift we encode values of register cells with sets 
$$X^1=\{x_{2},x_{3},\ldots,x_{20}\}, X^2=\{x_{3},x_{4},\ldots,x_{21}\},..., X^e=\{x_{e+1},x_{e+2},\ldots,x_{e+19}\}.$$ 
It is clear that $X^{out}=\{x_1,\ldots,x_e\}$. Thus the set of encoding variables for this program is $X=\{x_{1},x_{2},\ldots,x_{e+19}\}$, and the propositional encoding is the following set of Boolean formulae
\begin{equation*}
\begin{array}{l}
x_{20}\equiv x_1\oplus x_2 \oplus x_3 \oplus x_6\\
\ldots \\
x_{e+19}\equiv x_{e}\oplus x_{e+1} \oplus x_{e+2} \oplus x_{e+5}.\\
\end{array}
\end{equation*}
\end{example}

An important feature of the \textsc{Transalg} system is full support of conditional operators. Consider an arbitrary expression $\Phi(z_1,\ldots,z_k)$ of the TA language. Here $z_1,\ldots,z_k$ are program variables of the \texttt{bit} type. Suppose that they are linked with encoding variables $x_1,\ldots,x_k$. Also suppose that Boolean formula $\phi(x_1,\ldots,x_k)$ is obtained as a result of translation of an expression $\Phi(z_1,\ldots,z_k)$. Below we say that an expression $\Phi(z_1,\ldots,z_k)$ is associated with a Boolean formula $\phi(x_1,\ldots,x_k)$.

The BNF of conditional operator has the following form

\begin{verbatim}
<if_statement> := if (<expression>) <statement> [else <statement>]
\end{verbatim}
In this BNF \texttt{<expression>} is a predicate of a conditional operator. Let \texttt{<expression>} be $\Phi$. Suppose that expression $\Phi$ is associated with formula $\phi$. Denote by $\Delta_1$ and $\Delta_2$ some expressions of a TA-program, associated with Boolean formulae $\delta_1$ and $\delta_2$, respectively. Now suppose that program variable $z$ is the left operand in the assignment operator $z=\Delta_1$, performed in the first branch of the conditional operator and also the left operand of the assignment operator $z=\Delta_2$, performed in the second (else-) branch of the conditional operator. Also, suppose that at the previous translation step $z$ was linked with encoding variable $x$. During the translation of such conditional operator \textsc{Transalg} creates new encoding variable $x'$, links it with program variable $z$ and adds the following formula to the propositional encoding:
\begin{equation}
\label{eq1}
x'\equiv \phi\cdot\delta_1\vee\neg\phi\cdot\delta_2.
\end{equation}
If there is no assignment $z=\Delta_2$ in the else-branch, or if there is no else-branch, then formula \eqref{eq1} transforms into $x'\equiv\phi\cdot{\delta_1}\vee\neg\phi\cdot x$. Likewise if there is no assignment $z=\Delta_1$ in the first branch then formula \eqref{eq1} transforms into $ x'\equiv\phi\cdot x\vee\neg\phi\cdot \delta_2$.

Note that according to the BNF-definition any operator can be a branch of conditional operator. In particular, we can consider the construction of $n$ nested conditional operators. Without the loss of generality suppose that we have the following operator
\begin{equation*}
\begin{array}{l}
\text{if}\;\Phi_1\;z=\Delta_1\\
\text{else if}\;\Phi_2\;z=\Delta_2\\
\ldots \\
\text{else if}\;\Phi_n\;z=\Delta_n\\
\text{else}\;z=\Delta_{n+1}\\
\end{array}
\end{equation*}
Suppose that each expression $\Phi_j$, $j=1,\ldots,n$ is associated with Boolean formula $\phi_j$. Then during the translation of this operator \textsc{Transalg} will create new encoding variable $x'$ and add the following formula to the propositional encoding
\begin{equation}
\label{eq2}
x'\equiv\phi_1\cdot\delta_1\vee\neg\phi_1\cdot\phi_2\cdot\delta_2\vee\ldots\vee\neg\phi_1\cdot\neg\phi_2\cdot\ldots\cdot\phi_n\cdot\delta_n\vee\neg\phi_1\cdot\neg\phi_2\cdot\ldots\cdot\neg\phi_n\cdot\delta_{n+1}.
\end{equation}
Here Boolean formulae $\delta_j$, $j=1,\ldots,n+1$ are associated with expressions $\Delta_j$. Note that \eqref{eq2} follows from \eqref{eq1}.

\section{Encoding Cryptanalysis Problems with \textsc{Transalg}}
In recent years a number of papers about the application of SAT-solvers to solving cryptanalysis problems \cite{DBLP:conf/sat/EibachPV08,Mcdonald_attackingbivium,DBLP:conf/sat/MironovZ06,DBLP:conf/pact/SemenovZBP11,DBLP:conf/tools/Soos10,DBLP:conf/sat/SoosNC09} were published. We used \textsc{Transalg} to make propositional encodings for a number of cryptographic functions.


\begin{figure}
	\centering
	\includegraphics[width=12cm]{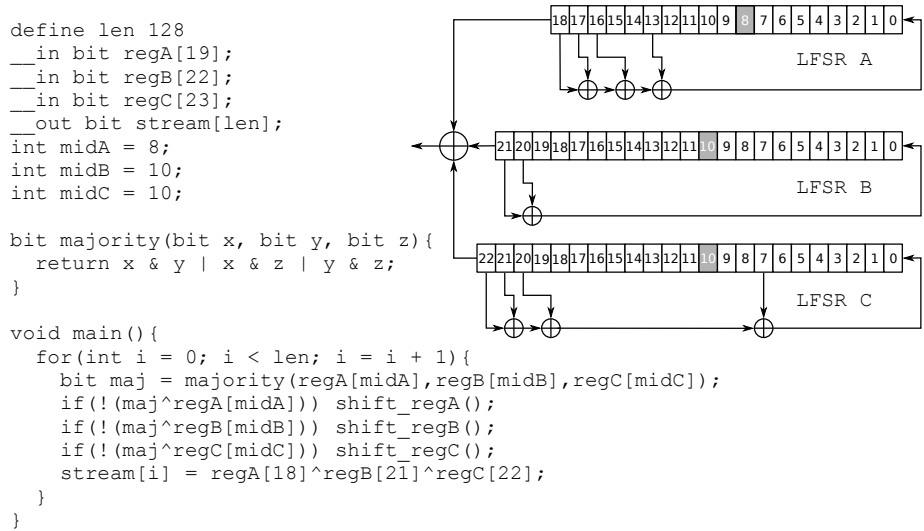}
	\caption{Fragment of the TA-program for the A5/1 keystream generator}
	\label{A5_1_ta}
\end{figure}
In fig. \ref{A5_1_ta} we show the fragment of the TA-program for the A5/1 generator outputting 128 bits of keystream. Here we suppose that functions \texttt{shift\_regA}, \texttt{shift\_regB} and \texttt{shift\_regC} are implemented in a similar way as \texttt{shift\_reg} function in fig. \ref{rslos_ta}. To translate Boolean formulae to CNF it is sufficient to use only Tseitin transformations \cite{Tseitin83}. For the TA-program in fig. \ref{A5_1_ta} this approach gives us the CNF with 41600 clauses over the set of 10816 variables.

However, in some cases the use of specialized Boolean optimization libraries makes it possible to significantly decrease the size of propositional encoding. In particular, the \textsc{Transalg} system uses the \textsc{Espresso}\footnote{http://embedded.eecs.berkeley.edu/pubs/downloads/espresso/index.htm} Boolean minimization library. In \textsc{Espresso} the Boolean formulas are minimized via the manipulation with their truth tables, that is why the complexity of the minimization procedure grows exponentially on the number of variables in the formula. In practice, it means that in order to spend reasonable time on minimization, it is best to minimize only formulas that contain limited number of variables. We use the following approach: if during the translation of the TA-program there arise the formulas over the set with more than $12$ variables then such formulas are divided into subformulas using the Tseitin transformations.

With the help of \textsc{Espresso} the size of the propositional encoding of the A5/1 keystream generator outputting 128 keystream bits was reduced to 39936 clauses over the set of 8768 variables. Note, that to obtain the encoding for the A5/1 generator with different size of keystream one only needs to change the value of \texttt{len} constant in the TA-program in fig. \ref{A5_1_ta} and repeat the translation. A5/1 encodings produced with \textsc{Transalg} were used to solve the problem of logical cryptanalysis of A5/1 in distributed computing environments \cite{DBLP:journals/corr/SemenovZ13,DBLP:conf/pact/SemenovZBP11}.

We also applied \textsc{Transalg} to encode the problem of cryptanalysis of the DES cipher to SAT. DES algorithm uses a lot of bit shifting and permutation operations. According to the translation concept used by \textsc{Transalg}, the system does not create new encoding variables during the translation of these operations. Propositional encodings of DES obtained with \textsc{Transalg} turned out to be significantly more compact than encodings from \cite{DBLP:journals/jar/MassacciM00}. For example, if we consider the DES algorithm that takes 1 block of plaintext (64 bits), the corresponding CNF obtained with \textsc{Transalg} has 26400 clauses over the set of 1912 variables. In \cite{DBLP:journals/jar/MassacciM00} CNF for the same problem has 61935 clauses over the set of 10336 variables.

The source code of the \textsc{Transalg}\footnote{https://gitlab.com/groups/transalg} system, the examples of TA-programs and the corresponding encodings\footnote{https://gitlab.com/groups/satencodings} are freely accessibly online. They include SAT encodings for problems of cryptanalysis of A5/1, A5/2, Bivium, Trivium, Geffe, Gifford, E0 and Grain keystream generators, DES algorithm, and also for hash algorithms MD4, MD5, SHA-1 and SHA-2.

\section{Related Work}

In recent years SAT community have developed many encoding techniques that can be applied to a wide class of combinatorial problems. A lot of references to key papers in this area can be found in \cite{DBLP:series/faia/Prestwich09}.

There is a number of systems for automated encoding of Constraint Satisfaction Problem (CSP) to SAT \cite{DBLP:conf/cpaior/HebrardOO10,DBLP:journals/tplp/MetodiC12,DBLP:conf/sat/SohTB13,tamura2008system}. Some methods and approaches to encoding of pseudoboolean constraints were described in \cite{DBLP:journals/jsat/EenS06}.

In \cite{DBLP:conf/tools/Soos10} there was presented the \textsc{Grain of Salt} tool that can be used to make SAT encodings for cryptanalysis of keystream generators. The \textsc{Grain of Salt} tool uses the special declarative language to describe keystream generators. This language was designed specifically to produce compact descriptions of configurations of shift registers. The \textsc{Transalg} system, on the contrary, was designed as a general tool, and it can be applied to encoding much wider class of functions than that covered by keystream generators. The ideology of \textsc{Transalg} is based on modern results in programming language theory and theory of symbolic execution. The most close analogue of the \textsc{Transalg} system is, apparently, the \textsc{URSA} system \cite{journals/lmcs/predrag}. In comparison to this tool, the distinctive feature of the \textsc{Transalg} system consists in ability to encode conditional operators.

The \textsc{Transalg} system showed high effectiveness in obtaining propositional encodings of various cryptographic functions. In particular, the SAT encodings of corresponding functions produced by \textsc{Transalg} are much more compact than known analogues.

\subsubsection*{Acknowledgments.} 
Authors thank Oleg Zaikin and Alexey Ignatiev for constructive feedback and helpful discussions. This work was partially supported by Russian Foundation for Basic Research, grants 14-07-00403a, 15-07-07891a, 14-07-31172mol\_a and by the President of Russian Federation grant for young scientists SP-3667.2013.5. 

\bibliographystyle{splncs03}
\bibliography{transalg_arx}

\end{document}